\pdfoutput=1

\documentclass[11pt]{article}
\usepackage[table,xcdraw]{xcolor}

\usepackage[preprint]{acl}
\usepackage{xcolor}
\usepackage{times}
\usepackage{latexsym}

\usepackage{booktabs}   
\usepackage{siunitx}    
\usepackage{array}      
\usepackage{multirow}   
\usepackage{arydshln}   
\usepackage{makecell}

\usepackage[normalem]{ulem}
\useunder{\uline}{\ul}{}
\usepackage{arydshln}

\usepackage{marvosym}
\usepackage{footmisc}
\usepackage{amsmath}

\usepackage{float}

\usepackage[T1]{fontenc}

\usepackage[utf8]{inputenc}

\usepackage{microtype}

\usepackage{inconsolata}

\usepackage{graphicx}

\title{DongbaMIE: A Multimodal Information Extraction Dataset for Evaluating Semantic Understanding of Dongba Pictograms}

\author{
 \textbf{Xiaojun Bi\textsuperscript{*,\Letter,1,2}},
 \textbf{Shuo Li\textsuperscript{*,3}},
 \textbf{Junyao Xing\textsuperscript{3}},
 \textbf{Ziyue Wang\textsuperscript{4}},
 \textbf{Fuwen Luo\textsuperscript{4}},
  \\
 \textbf{Weizheng Qiao\textsuperscript{1,2}},
 \textbf{Lu Han\textsuperscript{1,2}},
 \textbf{Ziwei Sun\textsuperscript{1,2}},
 \textbf{Peng Li\textsuperscript{5}},
 \textbf{Yang Liu\textsuperscript{4,5}},
\\
 \textsuperscript{1}College of Information and Engineering, Minzu University of China, Beijing, China \\
 \textsuperscript{2} \fontsize{11pt}{10pt}\selectfont Key Laboratory of Ethnic Language Intelligent Analysis and Security Governance of MOE, \\ \fontsize{11pt}{10pt}\selectfont Minzu University of China, Beijing, China \\
 \textsuperscript{3} \fontsize{11pt}{10pt}\selectfont College of Information and Communication Engineering, Harbin Engineering University, Harbin, China \\
 \textsuperscript{4} \fontsize{11pt}{10pt}\selectfont Dept. of Comp. Sci. \& Tech., Institute for AI, Tsinghua University, Beijing, China\\
  \textsuperscript{5} \fontsize{11pt}{10pt}\selectfont Institute for AI Industry Research (AIR), Tsinghua University, Beijing, China \\
   \texttt{\fontsize{11pt}{10pt}\selectfont \{bixiaojun, thinklis, 595587572\}@hrbeu.edu.cn}, \\
   \texttt{\fontsize{11pt}{10pt}\selectfont \{wangziyue22, lfw23\}@mails.tsinghua.edu.cn}\\
   \texttt{\fontsize{11pt}{10pt}\selectfont \{qiaoweizheng, 22400208, 22400215\}@muc.edu.cn}\\
  \texttt{\fontsize{11pt}{10pt}\selectfont lipeng@air.tsinghua.edu.cn, liuyang2011@tsinghua.edu.cn}
}

\begin{document}
\maketitle




\begin{abstract}
Dongba pictographic is the only pictographic script still in use in the world. Its pictorial ideographic features carry rich cultural and contextual information. However, due to the lack of relevant datasets, research on semantic understanding of Dongba hieroglyphs has progressed slowly. To this end, we constructed \textbf{DongbaMIE} - the first dataset focusing on multimodal information extraction of Dongba pictographs. The dataset consists of images of Dongba hieroglyphic characters and their corresponding semantic annotations in Chinese. It contains 23,530 sentence-level and 2,539 paragraph-level high-quality text-image pairs. The annotations cover four semantic dimensions: object, action, relation and attribute. Systematic evaluation of mainstream multimodal large language models shows that the models are difficult to perform information extraction of Dongba hieroglyphs efficiently under zero-shot and few-shot learning. Although supervised fine-tuning can improve the performance, accurate extraction of complex semantics is still a great challenge at present.\footnote{Our dataset can be obtained from: \url{https://github.com/thinklis/DongbaMIE}}

\end{abstract}

\section{Introduction}

\begin{figure}[t]
\centering
\includegraphics[width=\columnwidth]{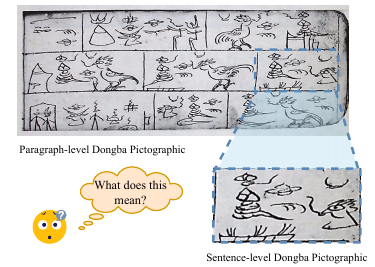}
  \caption{An example of Dongba pictographs from the DongbaMIE dataset.}
  \label{fig:1dataset}
\end{figure}

Dongba pictographic script is currently mainly used by the Naxi ethnic group in southwest China. As an important part of the cultural heritage of the Naxi ethnic group in China~\citep{LUO2023118865, xu2023digital}, ancient books written in Dongba pictographic script have been officially listed in the International Memory of the World Register by the United Nations Educational, Scientific and Cultural Organization (UNESCO). Figure~\ref{fig:1dataset} shows an image of Dongba pictographic script, which presents content composed of exquisite handwriting. It is recorded on local paper and mainly involves Dongba religion, Naxi history and lifestyle. However, there are severe challenges in the digital preservation and processing of Dongba pictographic script \citep{wang2011research}. First, since there are very few people who understand Dongba pictographic script, only a few Naxi priests (``Dongba'') can read this script today. This makes the relevant modern language annotation resources extremely scarce. Secondly, Dongba pictographic script lacks a standardized encoding system, and its grammatical structure is also significantly different from that of modern languages \citep{bi2024incomplete}. This makes it impossible to process it like regular text. These factors make it difficult to apply traditional natural language processing methods to the processing of Dongba pictographic scripts, and further aggravate the severe shortage of modern annotation corpus. Therefore, it is particularly important to construct a multimodal semantic understanding dataset based on Dongba pictograms, which will provide valuable resources for research and application in this field.

Recent advances in deep learning have provided new opportunities for endangered language processing \citep{sommerschield-etal-2023-machine, lu2024application, zhang-etal-2024-hire, zhang-etal-2022-nlp}. However, the understanding of contextual semantic information in Dongba pictographs has not been further studied. In addition, Dongba pictographs have unique linguistic phenomena. This includes the omission of pictographs in contextual sentences and the presence of polysemous characters with multiple meanings. These characteristics further increase the difficulty of deciphering and preserving these endangered texts through computation.

To improve the semantic understanding of Dongba hieroglyphs, we introduce a novel multimodal information extraction dataset, DongbaMIE, which is derived from ``The Annotated Collection of Naxi Dongba Manuscripts'' \citep{heannotated}. DongbaMIE contains 23,530 sentence-level and 2,539 paragraph-level images of precisely scanned manuscripts. It is also annotated with object, action, relationship and attribute information from its Chinese translation text. We evaluated multimodal large language models (MLLMs) at DongbaMIE. This includes zero-shot/few-shot performance of proprietary models such as GPT-4o \citep{hurst2024gpt}, and supervised fine-tuning (SFT) performance of open-source models (e.g., LLaVA-Next \citep{liu2024llavanext}, Qwen2-VL \citep{wang2024qwen2vlenhancingvisionlanguagemodels} ). Experiments show that current MLLMs struggle with this task.
 For example, the object extraction F1 of GPT-4o is only 1.60 under zero-shot. Despite the performance improvement of SFT, MLLMs show significant shortcomings in extracting complex semantics, especially in extracting relations and attributes. Our contributions are as follows:

\begin{itemize}
\item[$\bullet$] We introduce DongbaMIE, the first multimodal information extraction dataset for Dongba pictographs, which provides richly annotated image-text pairs covering four key semantic dimensions.

\item[$\bullet$] We systematically evaluate the mainstream MLLMs on DongbaMIE, assessing their performance in different settings such as zero-shot, few-shot, and SFT.

\item[$\bullet$] We find that MLLMs have significant limitations in performing multimodal information extraction tasks for Dongba pictograms, especially in complex semantic understanding.
\end{itemize}

\section{Related Work}
\subsection{Processing of Ancient Endangered Languages}

In recent years, the digitization of ancient endangered languages has gradually become an important research direction in NLP \citep{alp-2023-ancient, ml4al-2024-1}. These languages face a serious data scarcity problem. Their writing systems usually have diverse and complex morphological features. This makes it difficult to directly use traditional text processing methods \citep{ignat-etal-2022-ocr, Buoy-2023-Toward, Barney-2021-Towards}. With the development of technology, related research has also experienced a transition from basic digitization to more advanced language analysis tasks \citep{sommerschield-etal-2023-machine}. This includes semantic and sentiment analysis \citep{yoo-etal-2022-hue, sahala-etal-2020-automated, pavlopoulos-etal-2022-sentiment}, translation \citep{kang-etal-2021-restoring, yousef-etal-2022-automatic-translation, jin-etal-2023-morphological}, and decryption \citep{luo-etal-2021-deciphering, Daggumati-2018-Data-Mining}. The EvaLatin Challenge uses Perseus and LASLA corpora for Latin part-of-speech tagging and word form restoration \citep{sprugnoli-etal-2024-overview, sprugnoli-etal-2022-overview}.  \citet{yoo-etal-2022-hue} proposed a Transformer-based analysis of Chinese historical documents. \citet{jin-etal-2023-morphological} performed morphological and semantic evaluation on machine translation of ancient Chinese. However, the above research is still a text-centric approach. This makes it difficult to use it in ideographic writing systems with strong semantics. 

In addition, \citet{sahala-etal-2020-automated} proposed a character-level sequence-to-sequence model, which achieved the automatic transcription of Akkadian transliterated text for the first time. \citet{gordin-etal-2024-cured} proposed an OCR pipeline for digitizing cuneiform transliterated data. At the same time, \citet{cao-etal-2024-deep} proposed a method for translating ancient Egyptian hieroglyphs based on the Transformer model by combining the characteristics of speech and semantics. These methods mainly use the characteristics of speech and semantics to associate with the target language. Dongba hieroglyphs lack similar linguistic tool support due to their limited application and separation from the modern language system.

At present, the research on Dongba pictograms mainly focuses on the detection and recognition of a limited number of single characters \citep{LUO2023118865, ma2024stef}. There are also attempts to build a statistical translation model based on dependency structure \citep{gao2017chinese, gao2018chinese}. However, Dongba pictograms lack a unified encoding and no fixed grammatical phenomena. Previous methods have difficulty in obtaining contextual semantic information. In addition, due to the lack of a standard dataset, the language analysis task of Dongba pictograms is still in the initial exploration stage.

\subsection{Multimodal Semantic Analysis of Pictographic Characters}

In recent years, multimodal information extraction techniques have shown great potential in processing vision-language interaction tasks \citep{liu-etal-2019-graph, Sun_Zhang_Li_Lou_2024, Luo_2023_CVPR}. Existing studies have made significant progress in image-text alignment and understanding of historical documents and other pictographic symbols \citep{yang-etal-2023-histred, el2024handwritten, carlson-etal-2024-efficient} used contrastively trained visual encoders to model OCR as a character-level image retrieval problem. This achieved more accurate OCR results in historical documents. However, these methods mainly target standardized writing systems with clear encoding schemes. Recently, some studies have processed ideograms through novel multimodal methods. The OBI-BENCH benchmark \citep{chen2025obibenchlmmsaidstudy} evaluates large multimodal models on oracle bone script tasks (recognition, reconstruction, classification, retrieval, and interpretation). And LogogramNLP \citep{chen-etal-2024-logogramnlp} explores the learning of visual representations of ancient ideograms. These studies show that cultural heritage data can be unlocked by using multimodal processing methods.

While multimodal approaches have made progress in the identification of other historical scripts, they still struggle with the unique Dongba hieroglyphs. The lack of a standardized Dongba hieroglyphic dataset severely hampers the development and comparison of custom models.

\section{Constructing DongbaMIE Dataset}

\subsection{Data Collection}

The DongbaMIE dataset comes from the 100-volume ``Annotated Collection of Naxi Dongba Manuscripts'' \citep{heannotated}. This is a basic compilation of expertly verified Dongba hieroglyphic manuscripts and their Chinese translations. This comprehensive collection, which previously existed only in printed form, is a core resource for research on Dongba culture. Our current research focuses on digitizing the first ten volumes. This scope was determined based on a balanced representation of content and ensuring coverage of key Dongba language phenomena and characteristics. This was validated through experts in Dongba hieroglyphics. We obtained scanning authorization through the university library and worked with professional organizations. We specified a three-phase digitization program:

\begin{figure}[t]
  \includegraphics[width=\columnwidth]{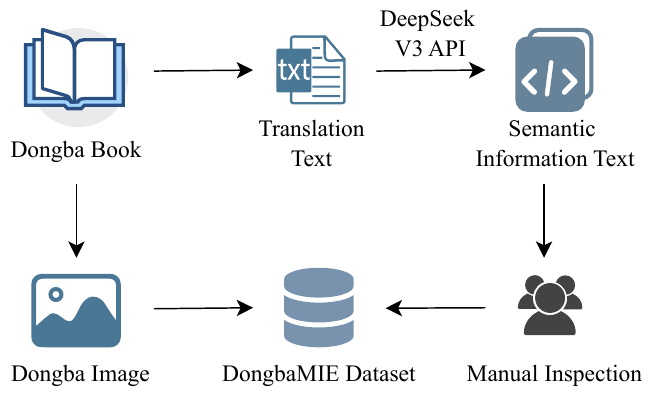} 
  \caption {The entire process of establishing the DongbaMIE dataset.}
  \label{fig:dataset_contruction}
\end{figure}


\begin{table*}
\centering
\small
\renewcommand{\arraystretch}{1.1}

\sisetup{
    group-separator={,},
    group-minimum-digits=4,
    detect-all
}

\begin{tabular}{
  l 
  S[table-format=5.0] 
  S[table-format=4.0] 
  S[table-format=5.0] S[table-format=5.0] 
  S[table-format=5.0] S[table-format=5.0] 
  S[table-format=5.0] S[table-format=5.0] 
  S[table-format=5.0] S[table-format=5.0] 
}
\toprule
\multirow{2}{*}{\textbf{Split}} &
  \multicolumn{1}{c}{\multirow{2}{*}{\textbf{Sentence}}} &
  \multicolumn{1}{c}{\multirow{2}{*}{\textbf{Paragraph}}} &
  \multicolumn{2}{c}{\textbf{Object}} &
  \multicolumn{2}{c}{\textbf{Action}} &
  \multicolumn{2}{c}{\textbf{Relation}} &
  \multicolumn{2}{c}{\textbf{Attribute}} \\
\cmidrule(lr){4-5} \cmidrule(lr){6-7} \cmidrule(lr){8-9} \cmidrule(lr){10-11}
& & & 
  \multicolumn{1}{c}{\textbf{Sent}} & \multicolumn{1}{c}{\textbf{Para}} &
  \multicolumn{1}{c}{\textbf{Sent}} & \multicolumn{1}{c}{\textbf{Para}} &
  \multicolumn{1}{c}{\textbf{Sent}} & \multicolumn{1}{c}{\textbf{Para}} &
  \multicolumn{1}{c}{\textbf{Sent}} & \multicolumn{1}{c}{\textbf{Para}} \\
\midrule
Train & 18824 & 2031 & 65264 & 52011 & 33658 & 24190 & 30165 & 29160 & 19754 & 19085 \\
Dev   & 2353  & 254  & 7966  & 6413  & 4144  & 2866  & 3720  & 3623  & 2653  & 2528 \\
Test  & 2353  & 254  & 8142  & 6716  & 4199  & 3052  & 3715  & 3974  & 2541  & 2497 \\
\hdashline %
Total \rule{0pt}{2.5ex} & 23530 & 2539 & 81372 & 65140 & 42001 & 30108 & 37600 & 36757 & 24948 & 24110 \\
\bottomrule
\end{tabular}
\caption{Statistics of the DongbaMIE dataset. ``Sent'' and ``Para'' denote sentences and paragraphs, respectively.}
\label{tab:dataset_statistics}
\end{table*}

\textbf{High-Fidelity Imaging and Preprocessing}: Images were acquired using professional-grade, high-resolution (e.g., 600 dpi) non-contact book scanners. Through pre-physical flattening, exposure and contrast optimization, and subsequent automatic paging and image enhancement processing, the JPG format image is ensured to retain the original handwriting details and spatial layout to the greatest extent.

\textbf{Structural Segmentation}: Leveraging the unique vertical separators in Dongba Pictograms as punctuation, we precisely segmented paragraph-level images into sentence-level images, while also retaining paragraph-level images to support multi-granularity analysis.

\textbf{Cross-Modal Alignment and Textual Correction}: Chinese translations within the images were initially extracted using OCR. However, all textual content is strictly manually proofread word by word. Trained annotators rectified errors by referencing the original manuscripts, ensuring the accuracy of the TXT-formatted text.

To ensure data quality, we established a multi-tiered Quality Assurance mechanism. This mechanism included real-time reviews by a professional team, systematic pre-service training for annotators, and the deep involvement of senior Dongba research experts throughout the process. These experts provided professional guidance, resolved ambiguous cases, and oversaw final quality control, guaranteeing the dataset's scholarly rigor.

\subsection{Annotation Scheme}

\begin{figure}[t]
\includegraphics[width=\columnwidth]{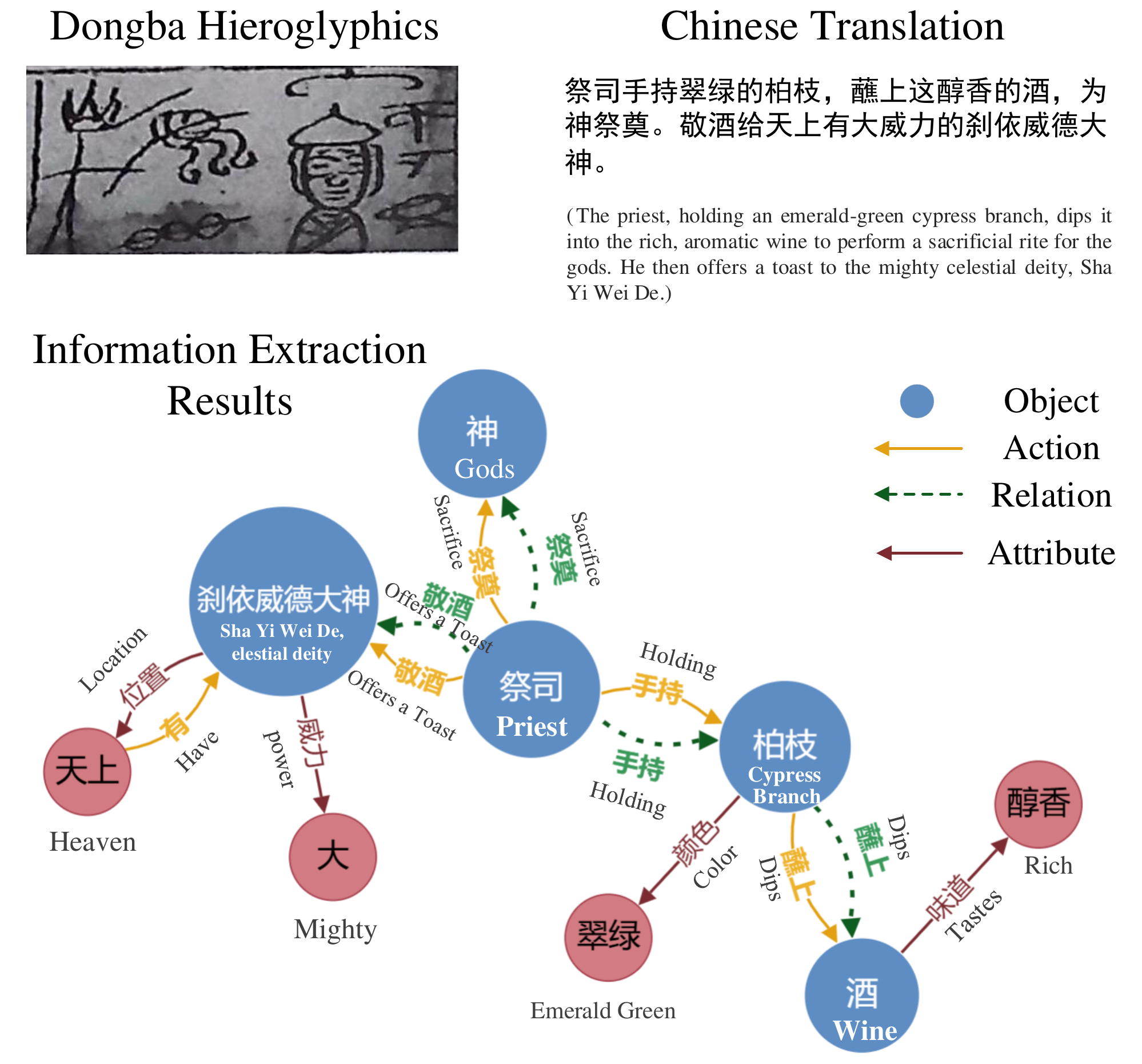}
  \caption{The image presents a semantic visualization yielded by the information extraction framework. Above this visualization, sentence-level Dongba pictographs are shown with their Chinese translations. English descriptions are provided solely for understanding; all annotations are originally in Chinese.}
  \label{fig:dongba_graph}
\end{figure}

We propose a multidimensional annotation framework which includes objects, relations, actions and attributes. The design of the framework is based on a systematic analysis of grammatical structures observed in Chinese translations of Dongba hieroglyphic. Our analysis reveals consistent correlations in which noun constituents typically denote objects, verb constituents express actions, prepositional structures encode relations, and modifiers describe attributes. Thus, this multidimensional structure effectively captures the core semantic elements of a sentence. Figure~\ref{fig:dongba_graph} provides an example detailing the Dongba hieroglyphs, their Chinese translation, and the corresponding information extraction results. Our multidimensional annotation framework contains four dimensions:


\textbf{Object:} They represent named entities and concepts derived from Chinese translations (e.g., ``priest'', ``wine''). These concepts are the key basic narrative units for understanding the theme.

\textbf{Action:} They represent events and activities, usually expressed as verbs (e.g., ``offer a ritual sacrifice'', ``offer a toast”). Actions constitute the narrative flow and its temporal causality.


\textbf{Relation:} They encode explicit connections and interactions between objects (e.g., ``priest-holding-cypress branches''). This dimension is essential for understanding the interactions between entities and maintaining contextual integrity.


\textbf{Attribute:} They specify descriptive features of an object or entity (e.g., ``wine-tastes-aromatic'', ``cypress branches-color-emerald green'') that add detail and depth to the semantic representation.



Overall, these four dimensions aim to holistically capture the multilayered semantics of Dongba hieroglyphs. Objects and their attributes constitute the core semantic entities, while actions and relationships describe the dynamic interactions in the narrative. The framework aims to balance the simplicity of annotation with comprehensive semantic coverage, providing a robust and scalable foundation for subsequent structured text analysis.

\begin{table*}[]
\small
\centering
\renewcommand{\arraystretch}{1.0} 
\setlength{\tabcolsep}{5pt}      

\sisetup{
    detect-all,
    round-mode=places,
    round-precision=2,
    group-separator={}, 
    table-align-text-pre=false, 
    table-align-text-post=false 
}

\newcolumntype{N}{S[table-format=2.2]}

\begin{tabular}{@{} l 
                   l 
                   c 
                   *{4}{N @{\ \ } N @{\ \ } N} 
                   @{}
                }
\toprule
\multicolumn{1}{c}{\multirow{2}{*}{\makecell[c]{\textbf{Model}}}} &
  \multicolumn{1}{c}{\multirow{2}{*}{\textbf{Level}}} &
  \multicolumn{1}{c}{\multirow{2}{*}{}} & 
  \multicolumn{3}{c}{\textbf{Object}} &
  \multicolumn{3}{c}{\textbf{Action}} &
  \multicolumn{3}{c}{\textbf{Relation}} &
  \multicolumn{3}{c}{\textbf{Attribute}} \\
\cmidrule(lr){4-6} \cmidrule(lr){7-9} \cmidrule(lr){10-12} \cmidrule(lr){13-15}
& & & 
  \multicolumn{1}{c}{P} & \multicolumn{1}{c}{R} & \multicolumn{1}{c}{F1} &
  \multicolumn{1}{c}{P} & \multicolumn{1}{c}{R} & \multicolumn{1}{c}{F1} &
  \multicolumn{1}{c}{P} & \multicolumn{1}{c}{R} & \multicolumn{1}{c}{F1} &
  \multicolumn{1}{c}{P} & \multicolumn{1}{c}{R} & \multicolumn{1}{c}{F1} \\
\midrule
\rule{0pt}{0ex}
\multirow{4}{*}{\makecell[c]{GPT-4o}} & \multirow{2}{*}{sent} & 0-shot & 1.73 & 1.80 & 1.60 & 0.34 & 0.39 & 0.32 & 0.00 & 0.00 & 0.00 & 0.00 & 0.00 & 0.00 \\
                                     &                       & 1-shot & 1.55 & 1.57 & 1.45 & 0.32 & 0.42 & 0.32 & 0.00 & 0.00 & 0.00 & 0.00 & 0.00 & 0.00 \\
                                     & \multirow{2}{*}{para} & 0-shot & 5.91 & 1.99 & 2.88 & 1.07 & 0.36 & 0.49 & 0.00 & 0.00 & 0.00 & 0.00 & 0.00 & 0.00 \\
                                     &                       & 1-shot & 6.85 & 2.18 & 3.16 & 1.28 & 0.61 & 0.76 & 0.00 & 0.00 & 0.00 & 0.00 & 0.00 & 0.00 \\
\hdashline
\rule{0pt}{2.5ex}
\multirow{4}{*}{\makecell[c]{Gemini-2.0}} & \multirow{2}{*}{sent} & 0-shot & 1.85 & 2.03 & 1.77 & 1.04 & 1.07 & 0.93 & 0.00 & 0.00 & 0.00 & 0.00 & 0.00 & 0.00 \\
                                         &                       & 1-shot & 1.61 & 1.79 & 1.55 & 1.12 & 1.33 & 1.09 & 0.00 & 0.00 & 0.00 & 0.00 & 0.00 & 0.00 \\
                                         & \multirow{2}{*}{para} & 0-shot & 6.39 & 2.04 & 2.91 & 4.47 & 1.44 & 2.08 & 0.00 & 0.00 & 0.00 & 0.00 & 0.00 & 0.00 \\
                                         &                       & 1-shot & 6.53 & 2.16 & 3.11 & 2.85 & 1.23 & 1.64 & 0.00 & 0.00 & 0.00 & 0.00 & 0.00 & 0.00 \\
\hdashline
\rule{0pt}{2.5ex}

\multirow{2}{*}{\makecell[c]{Qwen2-VL}} & sent & sft & 12.00 & 12.34 & 11.49 & 9.56 & 8.95 & 8.79 & 0.68 & 0.49 & 0.53 & 1.01 & 0.82 & 0.86 \\
                                       & para & sft & 4.00 & 5.84 & 4.43 & 6.54 & 6.01 & \colorbox[HTML]{CBCEFB}{\textbf{5.27}} & 0.10 & 0.19 & 0.11 & 0.99 & 0.90 & \colorbox[HTML]{CBCEFB}{\textbf{0.80}} \\
\hdashline
\rule{0pt}{2.5ex}
\multirow{2}{*}{\makecell[c]{CogVLM2}} & sent & sft & 7.97 & 7.49 & 7.22 & 1.76 & 1.51 & 1.50 & 0.06 & 0.06 & 0.06 & 0.00 & 0.00 & 0.00 \\
                                      & para & sft & 7.03 & 3.96 & \colorbox[HTML]{CBCEFB}{\textbf{4.71}} & 7.49 & 2.81 & 3.53 & 0.00 & 0.00 & 0.00 & 0.00 & 0.00 & 0.00 \\
\hdashline
\rule{0pt}{2.5ex}
\multirow{2}{*}{\makecell[c]{MiniCPM-V}}& sent & sft & 15.97 & 16.26 & 15.26 & 12.74 & 12.33 & 11.95 & 1.08 & 1.14 & \colorbox[HTML]{FFFFC7}{\textbf{1.00}} & 1.84 & 1.79 & \colorbox[HTML]{FFFFC7}{\textbf{1.69}} \\
                                        & para & sft & 3.93 & 5.69 & 4.31 & 5.28 & 5.73 & 4.83 & 0.10 & 0.20 & 0.13 & 0.39 & 0.71 & 0.48 \\
\hdashline
\rule{0pt}{2.5ex}
\multirow{2}{*}{\makecell[c]{LLava-Next}}& sent & sft & 17.06 & 17.27 & \colorbox[HTML]{FFFFC7}{\textbf{16.23}} & 14.10 & 12.88 & \colorbox[HTML]{FFFFC7}{\textbf{12.77}} & 1.03 & 0.78 & 0.84 & 1.78 & 1.55 & 1.54 \\
                                         & para & sft & 4.37 & 5.40 & 4.55 & 5.40 & 5.36 & 4.81 & 0.23 & 0.34 & \colorbox[HTML]{CBCEFB}{\textbf{0.25}} & 0.59 & 0.78 & 0.60 \\
\bottomrule
\end{tabular}
\caption{Model performance on the DongbaMIE dataset for extracting four semantic element types: Object, Action, Relation, and Attribute. We report Precision (P), Recall (R), and F1-score (F1) as percentages. Evaluations span sentence (Sent) and paragraph (Para) levels. For each element type, \colorbox[HTML]{FFFFC7}{yellow} highlighting marks the top sentence-level F1-score, while \colorbox[HTML]{CBCEFB}{purple} highlighting indicates the best paragraph-level F1-score.}
\label{tab:main_result}
\end{table*}

\subsection{Hybrid Annotation Pipeline}

To ensure the accuracy and reliability of semantic annotations in DongbaMIE, we implemented a rigorous hybrid annotation process. This process integrates automated pre-annotation using LLM with multi-stage manual verification and quality control by trained annotators. The overall workflow is shown in Figure~\ref{fig:dataset_contruction}.

\textbf{Automated Pre-annotation:} We initially employed the DeepSeek v3 API \citep{liu2024deepseek} for pre-annotation, extracting four key semantic dimensions—objects, actions, relations, and attributes—from the Chinese translations within the Annotated Collection of Naxi Dongba Manuscripts. This LLM-based step provided a consistent starting point, significantly reducing manual effort. An example extraction prompt is provided in Appendix \ref{sec:appendix_deepseek}.

\textbf{Manual Review and Refinement.}
Following pre-annotation, a team of 20 annotators conducted a comprehensive manual review and refinement. This team comprised 19 graduate students in Ethnic Minority Languages and Literature or Computer Science, and one lecturer specializing in Dongba pictograph research. Six members possessed prior experience in Dongba digitalization. All annotators received standardized training on the fundamentals of Dongba Pictograms, semantic dimension guidelines, workflow protocols, and the annotation software. All manual tasks were performed using a custom-developed web application, as shown in Appendix ~\ref{sec:appendix_web}, with annotation results automatically generated and stored in JSON format on the server. The focus of this phase was to correct errors and omissions from the LLM, aligning the annotations with expert linguistic understanding.

\textbf{Multi-level Quality Monitoring.} We adopted a three-stage quality control process to ensure consistent and accurate annotations. First, each instance was annotated independently by two annotators in a double-blind setting. Second, a third annotator reviewed and resolved any disagreements through arbitration. Third, two team members continuously monitored the process by randomly reviewing about 5\% of the data. When deviations were found, they provided targeted feedback and conducted retraining sessions.

We quantitatively assessed inter-annotator agreement (IAA) using Cohen's Kappa ($\kappa$) scores for all four semantic dimensions at the sentence level to evaluate the reliability of the manual annotations. As detailed in Appendix ~\ref{sec:appendix_IAA_Score}, the average $\kappa$ values indicate high consistency, which are 0.803 for the training set, 0.777 for the validation set, and 0.726 for the test set. The ``Action''  dimension showed almost perfect agreement ($\kappa$ \textgreater \ 0.93), while the ``Relation'' dimension, reflecting higher semantic complexity, exhibited moderate to substantial agreement. These IAA results confirm the high quality and reliability of our annotation process.

The DongbaMIE dataset is built upon rigorously validated semantic annotations. It provides a strong foundation for advancing research in multimodal information extraction, particularly for complex, low-resource languages like Dongba Pictograms. Detailed statistics are presented in Table~\ref{tab:dataset_statistics}.

\begin{table*}[]
\small
\centering
\renewcommand{\arraystretch}{1.0} 
\sisetup{
    detect-all,
    round-mode=places,
    round-precision=2,
    group-separator={} 
}

\newcolumntype{N}{S[table-format=2.2]}

\begin{tabular}{@{} l 
                   l 
                   l 
                   *{4}{N @{\ \ } N @{\ \ } N} 
                   @{}
                }
\toprule
\multicolumn{1}{c}{\multirow{2}{*}{\makecell[c]{\textbf{Model}}}} &
  \multicolumn{1}{c}{\multirow{2}{*}{\textbf{Level}}} &
  \multicolumn{1}{c}{\multirow{2}{*}{\textbf{Mode}}} & 
  \multicolumn{3}{c}{\textbf{Object}} &
  \multicolumn{3}{c}{\textbf{Action}} &
  \multicolumn{3}{c}{\textbf{Relation}} &
  \multicolumn{3}{c}{\textbf{Attribute}} \\
\cmidrule(lr){4-6} \cmidrule(lr){7-9} \cmidrule(lr){10-12} \cmidrule(lr){13-15}
& & & 
  \multicolumn{1}{c}{P} & \multicolumn{1}{c}{R} & \multicolumn{1}{c}{F1} &
  \multicolumn{1}{c}{P} & \multicolumn{1}{c}{R} & \multicolumn{1}{c}{F1} &
  \multicolumn{1}{c}{P} & \multicolumn{1}{c}{R} & \multicolumn{1}{c}{F1} &
  \multicolumn{1}{c}{P} & \multicolumn{1}{c}{R} & \multicolumn{1}{c}{F1} \\
\midrule
\rule{0pt}{0ex}
\multirow{4}{*}{\makecell[c]{GPT-4o}} & \multirow{2}{*}{sent} & single & 1.73 & 1.80 & 1.60 & 0.34 & 0.39 & 0.32 & 0.00 & 0.00 & 0.00 & 0.00 & 0.00 & 0.00 \\
                                     &                       & multi  & 1.80 & 1.45 & 1.46 & 0.21 & 0.11 & 0.14 & 0.00 & 0.00 & 0.00 & 0.00 & 0.00 & 0.00 \\
                                     & \multirow{2}{*}{para} & single & 5.91 & 1.99 & 2.88 & 1.07 & 0.36 & 0.49 & 0.00 & 0.00 & 0.00 & 0.00 & 0.00 & 0.00 \\
                                     &                       & multi  & 8.14 & 1.74 & 2.74 & 1.28 & 0.21 & 0.33 & 0.00 & 0.00 & 0.00 & 0.00 & 0.00 & 0.00 \\
\hdashline 
\rule{0pt}{2.5ex}
\multirow{4}{*}{\makecell[c]{Gemini-2.0}} & \multirow{2}{*}{sent} & single & 1.85 & 2.03 & 1.77 & 1.04 & 1.07 & 0.93 & 0.00 & 0.00 & 0.00 & 0.00 & 0.00 & 0.00 \\
                                         &                       & multi  & 1.48 & 1.83 & 1.50 & 0.30 & 0.21 & 0.23 & 0.00 & 0.00 & 0.00 & 0.00 & 0.00 & 0.00 \\
                                         & \multirow{2}{*}{para} & single & 6.39 & 2.04 & 2.91 & 4.47 & 1.44 & 2.08 & 0.00 & 0.00 & 0.00 & 0.00 & 0.00 & 0.00 \\
                                         &                       & multi  & 6.48 & 1.98 & 2.91 & 2.41 & 0.65 & 0.99 & 0.00 & 0.00 & 0.00 & 0.00 & 0.00 & 0.00 \\
\hdashline
\rule{0pt}{2.5ex}
\multirow{4}{*}{\makecell[c]{Qwen2-VL}} & \multirow{2}{*}{sent} & single & 12.00 & 12.34 & 11.49 & 9.56 & 8.95 & 8.79 & 0.68 & 0.49 & 0.53 & 1.01 & 0.82 & 0.86 \\
                                       &                       & multi  & 11.80 & 12.11 & 11.31 & 6.20 & 5.85 & 5.63 & 0.66 & 0.61 & 0.57 & 1.48 & 1.33 & 1.31 \\
                                       & \multirow{2}{*}{para} & single & 4.00 & 5.84 & 4.43 & 6.54 & 6.01 & 5.27 & 0.10 & 0.19 & 0.11 & 0.99 & 0.90 & 0.80 \\
                                       &                       & multi  & 4.65 & 6.73 & 5.14 & 5.18 & 6.84 & 5.14 & 0.03 & 0.05 & 0.04 & 0.16 & 0.27 & 0.19 \\
\hdashline
\rule{0pt}{2.5ex}
\multirow{4}{*}{\makecell[c]{CogVLM2}} & \multirow{2}{*}{sent} & single & 7.97 & 7.49 & 7.22 & 1.76 & 1.51 & 1.50 & 0.06 & 0.06 & 0.06 & 0.00 & 0.00 & 0.00 \\
                                      &                       & multi  & 7.75 & 8.63 & 7.65 & 2.70 & 2.52 & 2.40 & 0.67 & 0.62 & 0.58 & 0.13 & 0.17 & 0.14 \\
                                      & \multirow{2}{*}{para} & single & 7.03 & 3.96 & 4.71 & 7.49 & 2.81 & 3.53 & 0.00 & 0.00 & 0.00 & 0.00 & 0.00 & 0.00 \\
                                      &                       & multi  & 7.94 & 3.85 & 4.74 & 0.00 & 0.00 & 0.00 & 0.00 & 0.00 & 0.00 & 0.00 & 0.00 & 0.00 \\
\hdashline
\rule{0pt}{2.5ex}
\multirow{4}{*}{\makecell[c]{MiniCPM-V}}& \multirow{2}{*}{sent} & single & 15.97 & 16.26 & 15.26 & 12.74 & 12.33 & 11.95 & 1.08 & 1.14 & 1.00 & 1.84 & 1.79 & 1.69 \\ 
                                        &                       & multi  & 4.40 & 5.06 & 4.41 & 2.44 & 2.25 & 2.11 & 0.19 & 0.21 & 0.18 & 0.44 & 0.30 & 0.32 \\
                                        & \multirow{2}{*}{para} & single & 3.93 & 5.69 & 4.31 & 5.28 & 5.73 & 4.83 & 0.10 & 0.20 & 0.13 & 0.39 & 0.71 & 0.48 \\ 
                                        &                       & multi  & 3.94 & 5.82 & 4.40 & 5.41 & 6.61 & 5.22 & 0.05 & 0.11 & 0.07 & 0.13 & 0.19 & 0.13 \\
\hdashline
\rule{0pt}{2.5ex}
\multirow{4}{*}{\makecell[c]{LLava-Next}}& \multirow{2}{*}{sent} & single & 17.06 & 17.27 & 16.23 & 14.10 & 12.88 & 12.77 & 1.03 & 0.78 & 0.84 & 1.78 & 1.55 & 1.54 \\
                                         &                       & multi  & 16.00 & 16.44 & 15.32 & 9.30 & 9.15 & 8.76 & 1.12 & 1.19 & 1.04 & 1.71 & 1.79 & 1.62 \\
                                         & \multirow{2}{*}{para} & single & 4.37 & 5.40 & 4.55 & 5.40 & 5.36 & 4.81 & 0.23 & 0.34 & 0.25 & 0.59 & 0.78 & 0.60 \\ 
                                         &                       & multi  & 4.05 & 5.27 & 4.28 & 3.72 & 3.71 & 3.18 & 0.12 & 0.20 & 0.12 & 0.29 & 0.39 & 0.30 \\
\bottomrule
\end{tabular}
\caption{Comparison of the performance of MLLMs in extracting single semantic dimensions and multiple concurrent semantic dimensions from Dongba hieroglyphs. Results are analyzed at the sentence and paragraph level, where P, R and F1 denote precision, recall and F1 score, respectively.}
\label{tab:Multi_IE_result}
\end{table*}

\section{Methodology}

This section details our experimental methodology for evaluating MLLMs on the DongbaMIE dataset. We conducted zero-shot and one-shot evaluations for leading proprietary MLLMs, specifically GPT-4o \citep{hurst2024gpt} and Gemini-2.0 \footnote{\url{https://deepmind.google/technologies/gemini/}}, and Supervised Fine-Tuning for a diverse set of open-source MLLMs including LLava-Next \citep{liu2024llavanext}, MiniCPM-V-2.6 \citep{yao2024minicpmvgpt4vlevelmllm}, Qwen2-VL \citep{wang2024qwen2vlenhancingvisionlanguagemodels}, and CogVLM2 \citep{hong2024cogvlm2visuallanguagemodels}.

\subsection{Zero-shot and Few-shot Evaluation}
\label{subsec:4.1_zero_or_few_shot}

We evaluated GPT-4o and Gemini-2.0 via Visual Question Answering. In the zero-shot setting, models received only the Dongba image and a predefined prompt targeting the four semantic dimensions of objects, actions, relations, and attributes. For the one-shot setting, a single image-annotation pair was added as an in-context example. As a subset of few-shot, one-shot provides just one example. The specific prompt templates designed for these zero-shot and one-shot settings are visually presented in Appendix ~\ref{sec:appendix_prompt}.

\subsection{Open-Source MLLMs Supervised Fine-Tuning}

We fine-tuned LLava-Next, MiniCPM-V-2.6, Qwen2-VL, and CogVLM2 on DongbaMIE using instruction-image pairs derived from the dataset. Most models underwent full SFT, where all parameters were updated. For CogVLM2, we employed LoRA, a parameter-efficient fine-tuning method. All models were trained for 3 epochs, with the best-performing checkpoint on a development set selected for testing.

To specifically probe the role of visual representations in this domain, a topic further analyzed in Section \ref{sec:visual_feature_learning}, we applied two distinct SFT strategies to Qwen2-VL. The first was Full SFT, updating all parameters of both its vision and language modules. The second was LLM-only SFT, where the vision encoder was frozen, and only the LLM parameters were fine-tuned. This comparative setup aimed to isolate the contribution of learned, domain-specific visual features to extraction performance.

\subsection{Evaluation}

Model performance was assessed using Precision (P), Recall (R), and F1-score, calculated at both sentence and paragraph levels. Evaluations covered four semantic dimensions: objects, actions, relations, and attributes. At the same time, we evaluate the model in both single-dimension extraction and concurrent multi-semantic extraction to provide a comprehensive benchmark.

\section{Experimental Analysis}

\subsection{Overall Performance}

Extracting multimodal semantic information from the DongbaMIE dataset is challenging for current MLLMs, as shown in Table~\ref{tab:main_result}. Initial evaluations revealed significant limitations in proprietary models like GPT-4o and Gemini-2.0 using zero-shot and one-shot settings. For instance, GPT-4o's 0-shot sentence-level object extraction F1 was only 1.60. Both models were largely ineffective for relation and attribute extraction, achieving F1 scores of 0.00. When given paragraph-level input or a 1-shot prompt, these models achieved only minor F1 score increases for object and action extraction. Such improvements were far from overcoming their core performance deficiencies, indicating their struggle to accurately interpret the detailed semantics of Dongba pictograms.

SFT on open-source MLLMs led to varied degrees of performance improvement. At the sentence level, LLava-Next performed best, achieving top F1 scores of 16.23 for Object and 12.77 for Action extraction. MiniCPM-V-2.6 was also competitive, excelling with a sentence-level Attribute F1 score of 1.69 and achieving strong Object and Action F1 scores. While Qwen2-VL improved over zero-shot results, LLava-Next and MiniCPM-V-2.6 surpassed it. CogVLM2, using parameter-efficient fine-tuning, underperformed significantly, especially for relation and attribute extraction. This may indicate the necessity of full fine-tuning for this domain.

Across all evaluated models, two prominent findings emerge. First, sentence-level extraction consistently surpasses paragraph-level performance, indicating MLLMs' struggles with longer pictographic contexts. Second, relation and attribute extraction remain exceptionally challenging for all evaluated models. For these more complex tasks, even LLava-Next and MiniCPM-V-2.6 demonstrated only slight gains. Their F1 scores remained very low, generally 1.69 or less. This underscores fundamental limitations in current MLLMs' capabilities for complex reasoning and nuanced attribute recognition from pictographic inputs.


\begin{figure}[t]
  \includegraphics[width=\columnwidth]{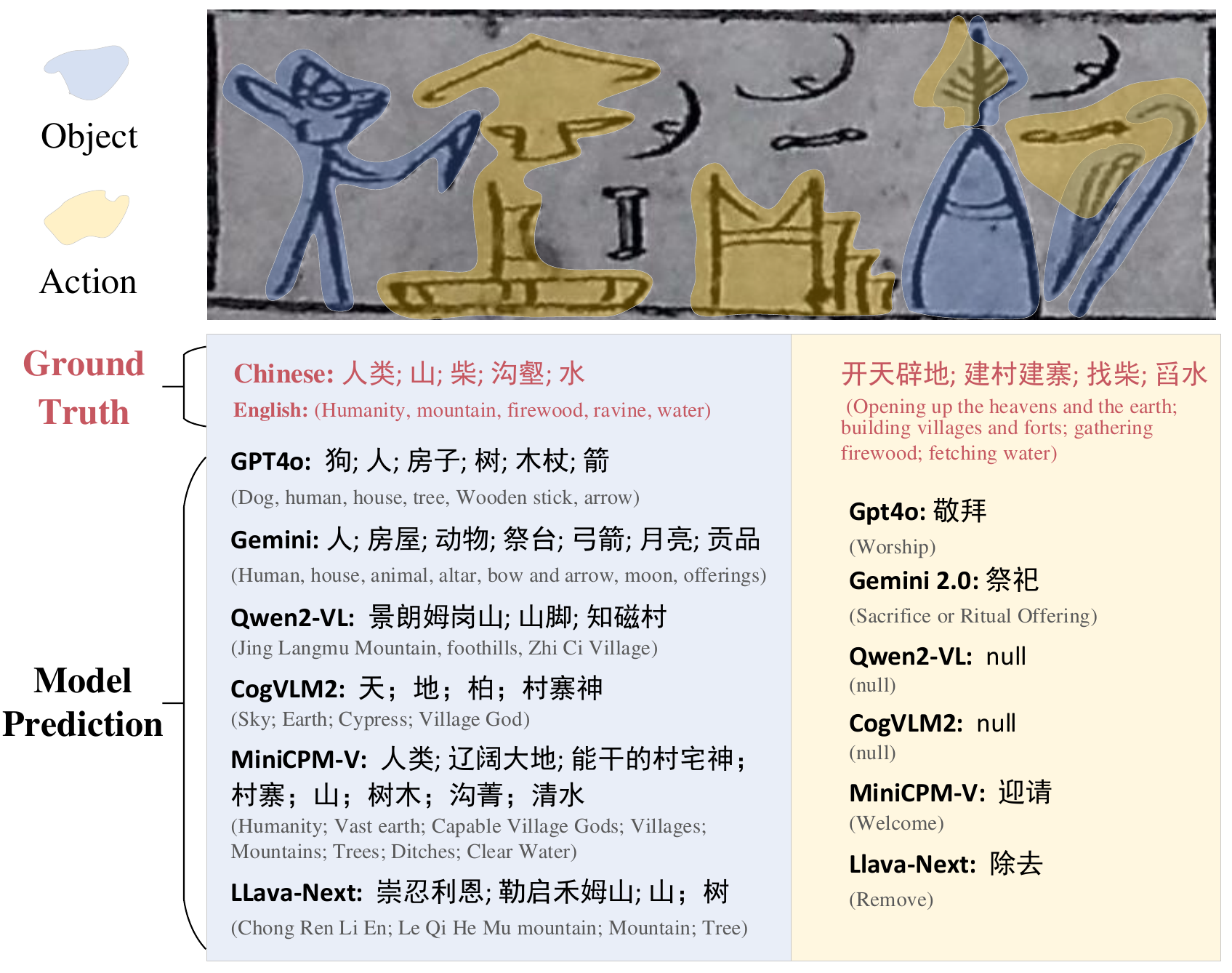}
  \caption{The image displays a Dongba pictograph with manually added annotations highlighting Objects (blue) and Actions (yellow). The text below shows multimodal semantic extraction results from six models, alongside ground truth labels. Prediction and Ground truth are shown in Chinese, with English notes for clarity.}
  \label{fig:case}
\end{figure}

\begin{table*}[]
\small
\centering
\renewcommand{\arraystretch}{1.0} 
\sisetup{
    detect-all,                 
    round-mode=places,          
    round-precision=2,          
    group-separator={}}         

\newcolumntype{N}{S[table-format=2.2]}

\begin{tabular}{@{} l l *{4}{N @{\ \ } N @{\ \ } N} @{}} 
\toprule
\multicolumn{1}{l}{\multirow{2}{*}{\makecell[c]{\textbf{Model}}}} & 
  \multicolumn{1}{c}{\multirow{2}{*}{\textbf{Level}}} & 
  \multicolumn{3}{c}{\textbf{Object}} &
  \multicolumn{3}{c}{\textbf{Action}} &
  \multicolumn{3}{c}{\textbf{Relation}} &
  \multicolumn{3}{c}{\textbf{Attribute}} \\
\cmidrule(lr){3-5} \cmidrule(lr){6-8} \cmidrule(lr){9-11} \cmidrule(lr){12-14} 
& & 
  \multicolumn{1}{c}{P} & \multicolumn{1}{c}{R} & \multicolumn{1}{c}{F1} &
  \multicolumn{1}{c}{P} & \multicolumn{1}{c}{R} & \multicolumn{1}{c}{F1} &
  \multicolumn{1}{c}{P} & \multicolumn{1}{c}{R} & \multicolumn{1}{c}{F1} &
  \multicolumn{1}{c}{P} & \multicolumn{1}{c}{R} & \multicolumn{1}{c}{F1} \\
\midrule
\multirow{2}{*}{\makecell[c]{Qwen2-VL\\(llm-sft)}} & 
  sent &
  12.00 & 12.34 & 11.49 &
  9.56 & 8.95 & 8.79 &
  0.68 & 0.49 & 0.53 &
  1.01 & 0.82 & 0.86 \\
 & para &
  4.00 & 5.84 & 4.43 &
  6.54 & 6.01 & 5.27 &
  0.10 & 0.19 & 0.11 &
  0.99 & 0.90 & 0.80 \\  \hdashline
\multirow{2}{*}{\makecell[c]{Qwen2-VL\\(full-sft)}} & 
  sent &  \rule{0pt}{2.5ex} 
  18.91 & 18.94 & 18.06 &
  17.26 & 15.47 & 15.58 &
  1.04 & 0.73 & 0.76 & 
  2.18 & 1.81 & 1.86 \\
 & para &
  6.47 & 7.12 & 6.43 &
  6.71 & 6.38 & 5.76 &
  0.12 & 0.05 & 0.07 &
  0.65 & 0.77 & 0.63 \\
\bottomrule
\end{tabular}
\caption{Fine-tuning of Qwen2-VL for semantic information extraction from Dongba hieroglyphs: comparison of LLM (llm-sft) model fine-tuning only vs. full model fine-tuning (full-sft, including visual module) in MLLMs. We report Precision (P), Recall (R), and F1-score (F1) as percentages.}
\label{tab:full-train}
\end{table*}

\subsection{Single vs. Multi-Semantic Extraction}

This section evaluates the multi-extraction performance of MLLMs in four semantic dimensions. We compare this with the single dimension extraction performance. Table~\ref{tab:Multi_IE_result} details the results.

Zero-shot models lack task-specific training. They typically struggle with concurrent multi-semantic extraction. For instance, GPT-4o’s F1 scores generally decreased across sentence and paragraph levels. Its sentence-level Object F1 score dropped from 1.60 to 1.46. Gemini’s performance also markedly declined, especially in Action extraction. Its sentence-level F1 score plummeted from 0.93 to 0.23. For both models, the already challenging relation and attribute extraction showed no improvement in multi-task settings. Their F1 scores remained zero. This underscores the inherent difficulty of concurrent complex semantic extraction without targeted training.

Models undergoing SFT showed varied performance in concurrent multi-semantic extraction. This task remained challenging for them. For CogVLM2, at the sentence level, the Object F1 score improved from 7.22 to 7.65, and the Action F1 score rose from 1.50 to 2.40. Previously unrecognized relation information was also extracted, achieving an F1 score of 0.58, along with attribute information, which reached an F1 score of 0.14. For performance at the paragraph level, there were varying degrees of increases or decreases. Other SFT models also displayed such inconsistencies. Improvements in some aspects or contexts were often offset by declines in others. Thus, SFT provides situational benefits but does not fully resolve concurrent extraction issues, frequently leading to improvements in some areas alongside performance drops in others.


\subsection{Impact of Visual Feature Learning}
\label{sec:visual_feature_learning}

We evaluated the impact of visual representation learning on model performance. To do this, we compared two Qwen2-VL setups: one with only the LLM fine-tuned, and another with full fine-tuning, including its visual module. Table \ref{tab:full-train} presents these results. The findings clearly demonstrate that full fine-tuning significantly enhances performance in Dongba pictograph multimodal semantic extraction. This improvement is particularly evident at the sentence level. For instance, the F1 score for Object extraction substantially increased from 11.49 to 18.06. Similarly, the Action extraction F1 score rose from 8.79 to 15.58. These results directly show that the model's ability to learn and optimize visual features specific to Dongba Pictograms is crucial for accurately understanding its pictographic semantics. Conversely, limited visual feature extraction capabilities are a key factor contributing to performance bottlenecks.

Full fine-tuning also yielded performance gains at the paragraph level. For example, Object F1 improved from 4.43 to 6.43. However, the extent of this improvement was less pronounced than at the sentence level. This observation may suggest that the contribution of high-quality visual features diminishes in longer textual contexts. Alternatively, current model architectures might still face limitations in effectively modeling long-range visual dependencies. Overall, these findings underscore the critical importance of targeted enhancements in visual feature learning for processing pictographic scripts like Dongba. They also provide clear directions for future work aimed at improving model capabilities.

\subsection{Error Analysis and Future Directions}

This section presents a qualitative case study based on the Dongba pictograph in Figure~\ref{fig:case}. It highlights key challenges faced by MLLMs in interpreting Dongba hieroglyphs. The analysis focuses on extracting objects and actions, which are essential for narrative understanding in hieroglyphic scripts.

Many models, including Gemini 2.0 and CogVLM2, failed to recognize core environmental objects such as ``mountain'' and ``water.'' Others, like Qwen2-VL and Llava-Next, replaced generic symbols with overly specific named entities, such as ``Le Qi He Mu mountain.'' MiniCPM-V showed better performance in extracting key objects.

Interpreting symbolic features proved problematic. This included erroneous segmentation of unified entities (GPT-4o) and overly abstract readings of potential human figures as ``village god'' (CogVLM2, MiniCPM-V).

Action extraction was acutely challenging. Models predominantly returned inaccurate, abstract labels (Gemini 2.0: ``sacrifice''; GPT-4o: ``worship''; MiniCPM-V: ``welcome''; Llava-Next: ``remove'') or no output (Qwen2-VL, CogVLM2), failing to identify specific activities such as ``gathering firewood.''

These challenges reveal the complexity of Dongba hieroglyphics. They highlight the urgent need for models with greater visual contextual understanding.


\section{Conclusion}

In this study, we introduced DongbaMIE, the first multimodal information extraction dataset for Dongba pictographs. It aims to advance the processing of endangered scripts. By evaluating against state-of-the-art closed-source and open-source MLLMs, we found the limitations of zero-shot and few-shot methods, while demonstrating the potential of supervised fine-tuning. Despite this, current models still struggle to understand the complexity of Dongba pictographs. In the future, we will focus on expanding the dataset annotations with more fine-grained annotations. At the same time, we will improve the multimodal representation to further enhance the ability to extract information from Dongba pictographs. Our dataset will be released soon after the acceptance of the paper. This is intended to facilitate reproducible research in the field of endangered text conservation.

\section*{Limitations}

Despite the innovative multimodal dataset presented in this study, there are still several limitations that need to be addressed. First, due to the limitations of current multimodal models in handling low-resource languages and pictographic scripts, these models still face significant challenges in accurately extracting semantic information from Dongba pictographs. In particular, the model performance remains limited when interpreting the actions, relations, and attributes conveyed by Dongba symbols. We believe that with the continued advancement of MLLMs, future models will be better equipped to handle complex image and semantic information, thereby improving the semantic understanding and information extraction of Dongba pictographs. Second, although the dataset covers multiple image levels and four semantic dimensions, we plan to expand the dataset in future work to explore more fine-grained symbol and semantic annotations, such as context-based semantic understanding of individual Dongba pictograph characters. Additionally, integrating data from other modalities, such as audio, video, or depth images, could provide the model with more contextual information, thus enhancing overall multimodal understanding capabilities.

\bibliography{custom}

\appendix

\clearpage
\twocolumn

\section{Data Annotation and Review Web Application}
\label{sec:appendix_web}

\begin{figure*}[!h]
\centering
  \includegraphics[width=\textwidth]{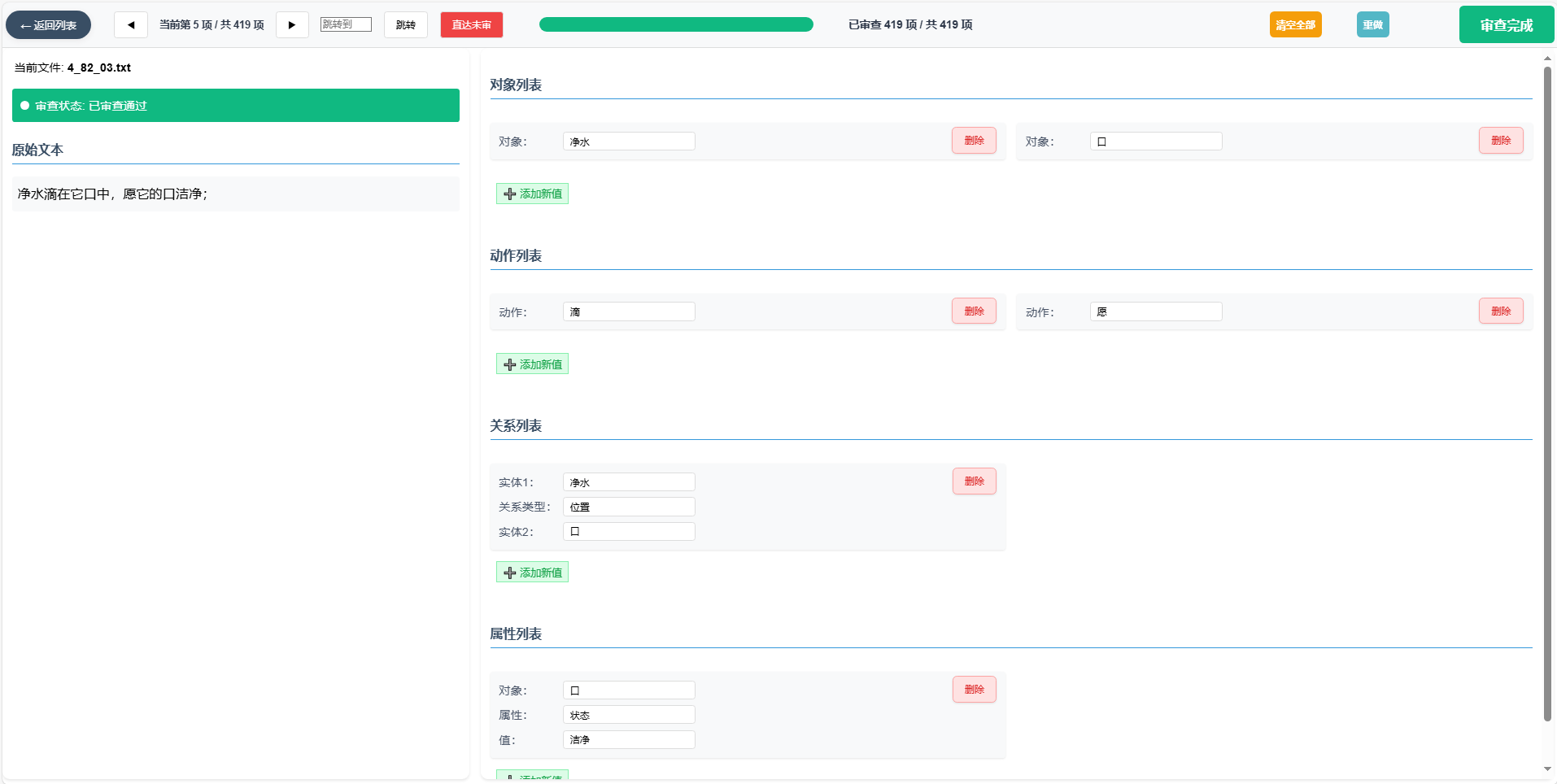}
  \caption{This is a web application page for manual review of Dongba pictographic information extraction.}
    \label{fig:appendix_web2}
\end{figure*}

We employed a custom-developed web application (Figure~\ref{fig:appendix_web2}) to manually review and refine pre-annotated Dongba pictograph data. This application standardizes the annotation workflow and improves data quality. It provides a visual interface for Dongba pictographs and their multimodal information. The application enables straightforward editing of pre-annotations to correct machine-generated errors and omissions. It also guides users according to established annotation protocols and automatically stores reviewed results in JSON format on the server-side.

\section{Inter-Annotator Agreement Scores}
\label{sec:appendix_IAA_Score}

To ensure the quality and reliability of our manual annotations, we conducted a quantitative evaluation of Inter-Annotator Agreement (IAA) at the sentence level. This evaluation covered all four core semantic dimensions: Action, Object, Relation, and Attribute. We employed Cohen's Kappa ($\kappa$) coefficient \citep{1960Coefficient} as the evaluation metric. Cohen's Kappa is widely used to measure agreement between annotators, accounting for agreement that could occur by chance \citep{mchugh2012interrater}. The IAA results are presented in Table \ref{tab:IAA_score}.

\textbf{Overall Agreement:} The average $\kappa$ values are 0.803 for the training set, 0.777 for the development set, and 0.726 for the test set. According to the standard interpretation of Kappa values \citep{landis1977measurement}, these scores indicate substantial to almost perfect agreement. This suggests a high overall reliability for our manual annotations.

\textbf{Dimensional Differences:} The Action dimension exhibits very high agreement across all datasets. Its $\kappa$ values consistently exceed 0.93 (Train: 0.981, Dev: 0.976, Test: 0.935), signifying almost perfect agreement. This indicates a strong consensus among annotators in identifying actions.
In contrast, the Relation dimension shows lower $\kappa$ values (Train: 0.671, Dev: 0.626, Test: 0.542). These scores fall within the moderate to substantial agreement range. This lower agreement likely reflects the inherent semantic complexity and subjectivity of the Relation dimension. Annotators may face more challenges in interpreting and labeling complex relationships, leading to more divergence. Nevertheless, these values still indicate an acceptable quality of annotation for this dimension.
The Object and Attribute dimensions also demonstrate good agreement. For instance, in the training set, the $\kappa$ values for Object (0.734) and Attribute (0.825) suggest consistent judgments by annotators.

\textbf{Cross-Dataset Comparison:} The average IAA for the training set (0.803) is slightly higher than those for the dev (0.777) and test (0.726) sets. These minor variations are normal. They may reflect subtle differences in data distribution or slight fluctuations in annotator performance across batches. However, the overall trend indicates that a high level of annotation quality was maintained across all datasets.

\begin{figure}[t]
\includegraphics[width=\columnwidth]{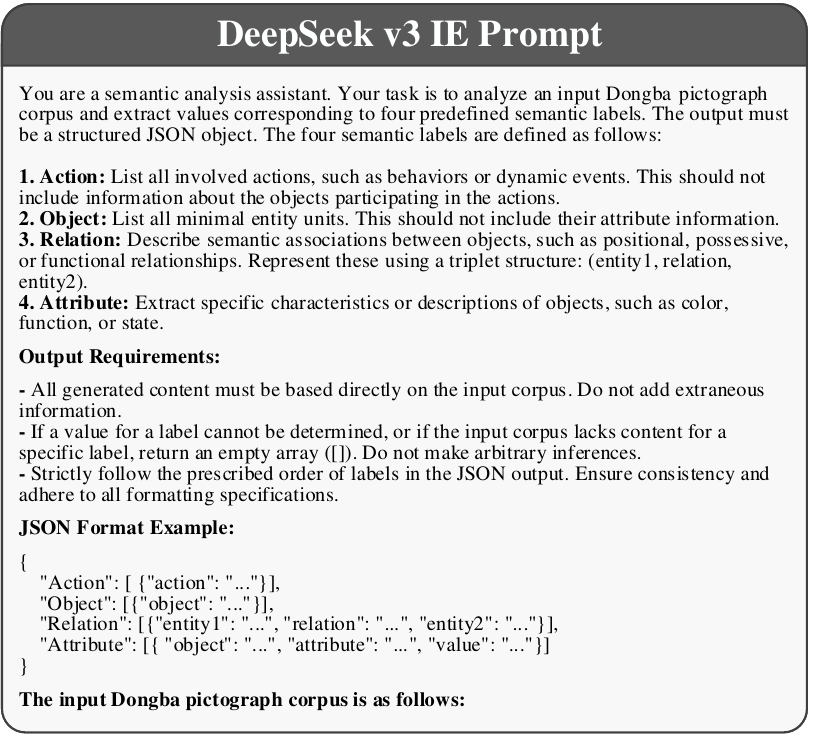}
  \caption{DeepSeek v3 IE Prompt template automates multi-label semantic analysis of Dongba Pictograms, outputting actions, objects, relations, and attributes in structured JSON.}
  \label{fig:deepseek_v3_IE_prompt}
\end{figure}

\begin{table*}[]
\centering
\begin{tabular}{@{}ccccccc@{}}
\toprule
\multirow{2}{*}{\textbf{Split}} &
  \multirow{2}{*}{\textbf{Sentence}} &
  \multirow{2}{*}{\textbf{Action}} &
  \multirow{2}{*}{\textbf{Object}} &
  \multirow{2}{*}{\textbf{Relation}} &
  \multirow{2}{*}{\textbf{Attribute}} &
  \multirow{2}{*}{\textbf{Avg.}} \\
      &       &       &       &       &       &       \\ \midrule
Train & 18824 & 0.981 & 0.734 & 0.671 & 0.825 & 0.803 \\
Dev   & 2,353 & 0.976 & 0.751 & 0.626 & 0.753 & 0.777 \\
Test  & 2,353 & 0.935 & 0.693 & 0.542 & 0.733 & 0.726 \\ \bottomrule
\end{tabular}
\caption{Cohen's Kappa ($\kappa$) Inter-Annotator Agreement (IAA) scores for each dataset (Train, Dev, Test), covering sentence counts, scores for each semantic dimension (Action, Object, Relation, Attribute), and average ($\text{Avg.}$) $\kappa$ values.}
  \label{tab:IAA_score}
\end{table*}

\section{Example Prompt for Automated Pre-annotation}
\label{sec:appendix_deepseek}

During automated pre-annotation, we utilized the DeepSeek v3 LLM API to extract key semantic dimensions from Chinese translations in the Naxi Dongba manuscript annotation corpus. Figure ~\ref{fig:deepseek_v3_IE_prompt} shows an example prompt designed to guide the model in this information extraction task. This prompt directs the model to identify and extract predefined semantic elements: entities, actions, relations, and attributes.

Specific guidelines within the prompt addressed nuanced aspects, such as defining the scope of actions or the contextual nature of entity attributes. To streamline downstream processing and support subsequent review by human annotators, the prompt requested output in a structured format (e.g., JSON). The prompt was carefully constructed with the goal of maximizing the accuracy and completeness of the semantic information captured from the translations.

\begin{figure}[t]
  \includegraphics[width=\columnwidth]{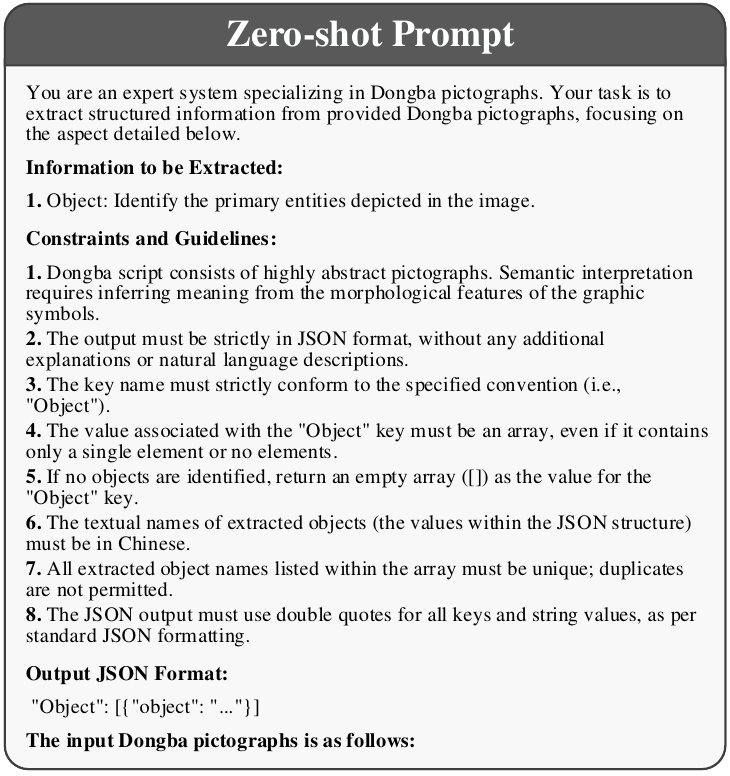}
  \caption{Zero-shot prompt template for MLLMs (e.g., GPT-4o, Gemini) to extract four semantic types from Dongba hieroglyphs.}
  \label{fig:zero_shot_prompt}
\end{figure}

\begin{figure}[t]
  \includegraphics[width=\columnwidth]{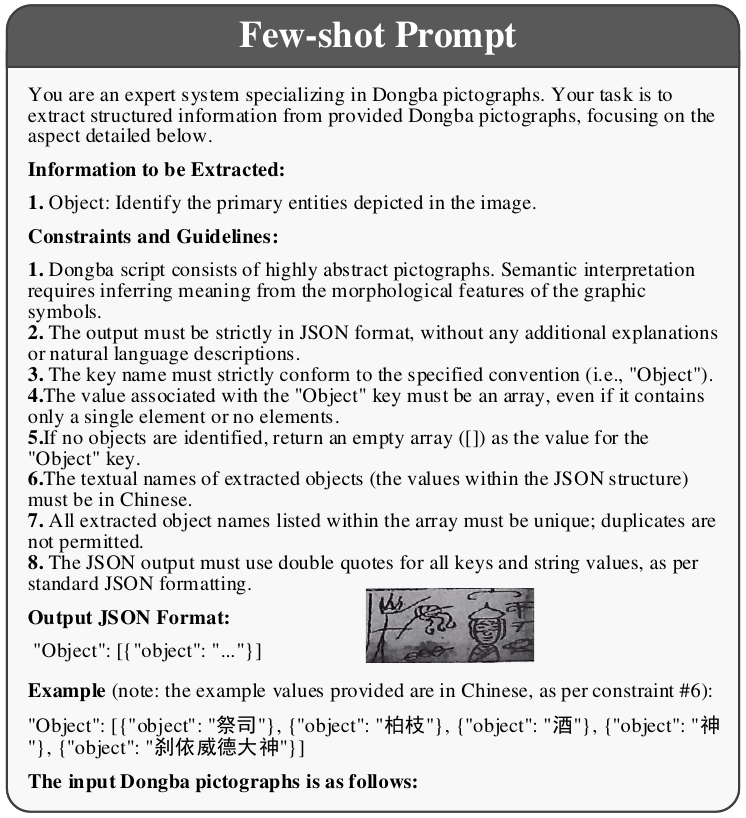}
  \caption{Few-shot prompt template for MLLMs (e.g., GPT-4o, Gemini) to extract four semantic types from Dongba pictographs.}
  \label{fig:few_shot_prompt}
\end{figure}

\section{Prompt Templates}
\label{sec:appendix_prompt}


This appendix presents prompt templates used to evaluate two proprietary MLLMs—GPT-4o and Gemini 2.0—on Dongba Pictograms information extraction (Section ~\ref{subsec:4.1_zero_or_few_shot}). These templates are presented in English for clarity. The operational versions used in our evaluations were authored and deployed in Chinese.

Figure~\ref{fig:zero_shot_prompt} shows the template for zero-shot semantic information extraction from Dongba pictographic. Applied to MLLMs like GPT-4o and Gemini 2.0, this prompt directs the model to identify and structure multiple semantic categories (e.g., actions, objects, relations, attributes). This process relies on instructions alone, without in-context examples.

Figure~\ref{fig:few_shot_prompt} illustrates the ``Few-shot Prompt'' template for one-shot (an instance of few-shot learning) diverse semantic extraction from Dongba pictographic images. This template includes an in-prompt example with a placeholder for an image-annotation pair. This guides the model to identify semantic elements (e.g., actions, objects, relations, attributes) from visual input.

\clearpage
\onecolumn

\end{document}